\documentclass{article}

\usepackage[preprint]{nips_2018}

\usepackage{natbib}
\usepackage[utf8]{inputenc} 
\usepackage[T1]{fontenc}    
\usepackage{hyperref}       
\usepackage{url}            
\usepackage{booktabs, multirow}       
\usepackage{amsfonts}       
\usepackage{nicefrac}       
\usepackage{microtype}      

\usepackage{graphicx}
\usepackage{epstopdf}

\usepackage{latexsym}
\usepackage{amsmath}

\usepackage{caption}
\usepackage[colorinlistoftodos]{todonotes}

\newcommand\blfootnote[1]{%
  \begingroup
  \renewcommand\thefootnote{}\footnote{#1}%
  \addtocounter{footnote}{-1}%
  \endgroup
}
\title{Interpreting Models by Allowing to Ask}

\author{
  Sungmin Kang$^\dagger$\\
  KAIST\\
  \texttt{stuatlittle@kaist.ac.kr}\\
  \And
  David Keetae Park \\
  Korea University\\
  \texttt{heykeetae@korea.ac.kr} \\
  \And
  Jaehyuk Chang\\
  NAVER WEBTOON Corp.\\
  \texttt{jaehyuk.chang@webtoonscorp.com}\\
  \And
   Jaegul Choo\\
  Korea University\\
  \texttt{jchoo@korea.ac.kr}\\
}

\begin{document}

\maketitle

\begin{abstract} 
	Questions convey information about the questioner, namely what one does not know. In this paper, we propose a novel approach to allow a learning agent to ask what it considers as tricky to predict, in the course of producing a final output. By analyzing when and what it asks, we can make our model more transparent and interpretable. We first develop this idea to propose a general framework of deep neural networks that can ask questions, which we call \textit{asking networks}. A specific architecture and training process for an asking network is proposed for the task of colorization, which is an exemplar one-to-many task and thus a task where asking questions is helpful in performing the task accurately. Our results show that the model learns to generate meaningful questions, asks difficult questions first, and utilizes the provided hint more efficiently than baseline models. We conclude that the proposed asking framework makes the learning agent reveal its weaknesses, which poses a promising new direction in developing interpretable and interactive models.
    
\end{abstract}

\section{Introduction} 

\blfootnote{\noindent$^\dagger$ This work was done as part of remote internship at NAVER WEBTOON Corp.} Questions play a crucial role in communication and learning. Not only do the questions of a curious child help her learn facts about the world, but also such questions help to understand what she learnt and what she did not. In this paper, we attempt to explore the merit of having questions from the perspective of observers, rather than learners. For instance, if the learning agent asks, ``what is in the box?'', and the answer is ``a cat'', we focus our attention on the new information that the agent did not know what is present in the box.

  Questions help us understand the information state of the model. Allowing models to ask questions thus brings a positive impact on the interpretability of a learning system. We explicitly explore the asking ability of a learning agent for a better understanding of the decisions it makes. Our research hypothesis is thus:

\textit{Allowing a learning agent to ask questions makes its inner-workings more interpretable to observers.}

To demonstrate our hypothesis, we propose a class of neural networks that can ask questions, which we call \textit{asking networks}. Following our asking paradigm an agent learns a task while iteratively asking questions and being guided by pre-defined answering rules. The answers should be designed to be beneficial for the agent to predict the label, so that in an attempt to reduce the loss, the agent would ask questions and take the answers in return. We focus on those questions the agent generates because they can be used as a means of a communication window. 




Among possible applications, we test our asking paradigm on deep automatic colorization, a field of conditional image generation in which the model performs colorization given a grayscale image or outline drawing. Deep colorization is inherently multimodal~\cite{zhang2017stackgan, zhang2017real}, meaning that there are multiple plausible outputs given the same input. The one-to-many ambiguity of deep colorization has been partially addressed by using an autoregressive method~\cite{guadarrama2017pixcolor} or by incorporating user priors and interactions~\cite{iizuka2016let,zhang2017real}. We instead design an asking network that asks for ground-truth colors in such circumstances where model output may vary.

Concretely, we allow our model to ask for the color of a model-chosen region of an image. Then, the single color answer, which is computed as the average ground-truth color of the corresponding region, is given to the model, and based on this answer the model performs colorization. We design our network in a recurrent setting so that the model asks questions one by one and updates the colorization output sequentially. 

\begin{figure}[t]
	\centering
    \includegraphics[width=\textwidth]{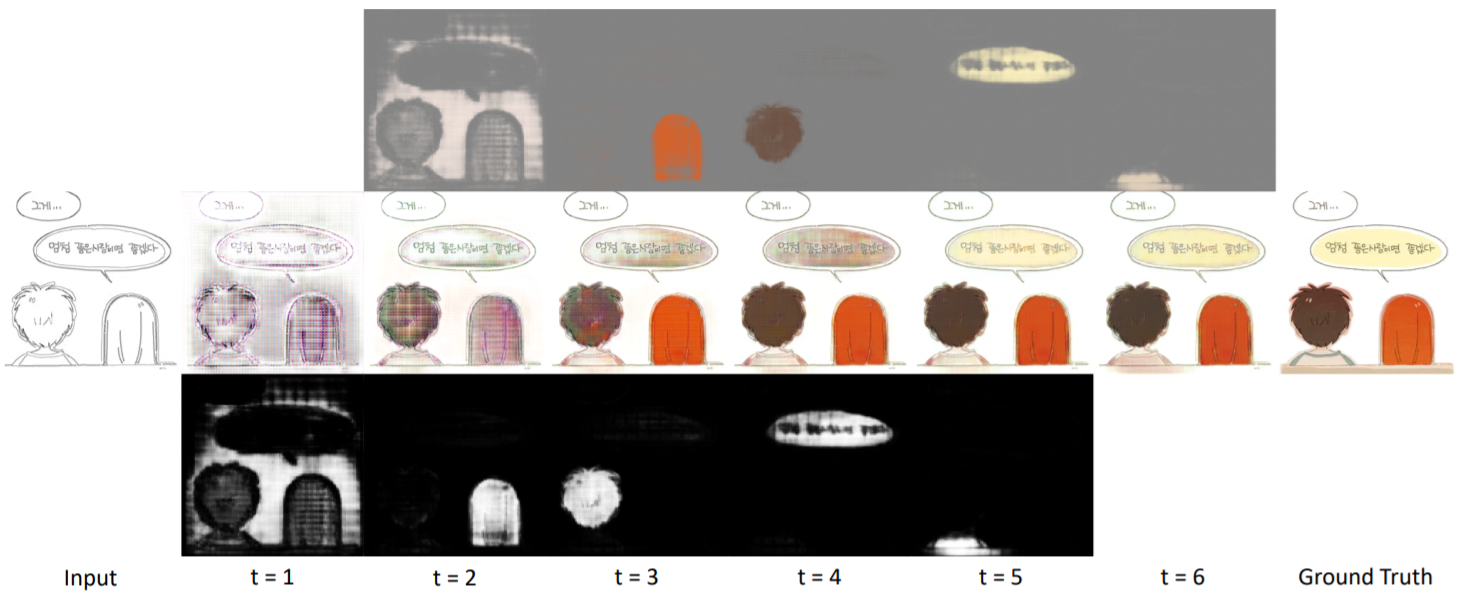}
    \caption{An example question-answer ambiguity reduction process as proposed by our model, showing a colorization process over multiple steps. The left and right extremes are the input image and the ground truth colorization, respectively; The middle row shows the intermediate colorization results, the bottom row shows the question asked by the model, represented as a pixel map, and the top row shows the answer to the question in the previous timestep. \copyright  Omyo.}
    \label{intro_img}
\end{figure}

Fig.~\ref{intro_img} illustrates an example of this process. Quantitative analyses show that our model learns to ask carefully thought-out questions to utilize the provided answer towards the maximum improvement of the loss function for colorization. Interestingly, the first question is shown to be the most effective for reducing the loss. The quantitative analysis on the VOC~\cite{pascal-voc-2012}, an image dataset with labeled class segmentation, also shows that our model learns semantically meaningful segmentations to be colorized as a single color and asks questions based on the learned segmentations. 

We complete our discussion by validating that questions generated by our model justify our initial hypothesis that questions help us interpret the learner. To summarize, the contribution of this paper is twofold. First, we introduce the new class of asking networks that are capable of interpretable modeling. Second, we develop an exemplary asking network in the deep colorization domain to show the potential application of our proposed approach and support our research hypothesis. 

\section{Related Work}


\paragraph{Interpretable deep learning}
Despite the rapid growth in the field of deep neural networks, due to their often unexplainable output and black-box nature there have been calls for methods to disclose the inner-workings of these complicated models. Interpretable models have been studied in different areas, such as image generation~\cite{chen2016infogan}, speech generation~\cite{hsu2017unsupervised} and text grounding for images~\cite{yeh2017interpretable}.       Many approaches to interpret the hidden features of convolutional neural networks rely on visualization, either directly visualizing activation maps~\cite{erhan2009visualizing,zeiler2014visualizing} or revealing the subparts of an image that are responsible for the predicted outputs~\cite{lu2017knowing,zhou2016learning}. These visualization methods are powerful tools for post-hoc analysis of a model~\cite{lipton2016mythos}, but they hardly convey information about the inner-workings of a model, i.e. which step of prediction attributes to such predictions. On the other hand, in our proposed asking paradigm, we explicitly stimulate models to expose weak parts, so that observers can interpret a particular stage of model behavior. 

\paragraph{Deep colorization}
Deep automatic colorization has been gaining significant attention owing to recent advances in deep learning based image processing and encoding methods~\cite{iizuka2016let,zhang2016colorful}. Most deep colorization models are composed of the U-net~\cite{ronneberger2015u} architecture, a popular encoder-decoder framework which has been adopted by many deep colorization studies~\cite{frans2017outline,chang2015palette}. Among practical applications of deep colorization is outline drawing colorization. Colorizing line sketches could mitigate laborious and repetitive work~\cite{frans2017outline}, because some shapes and characters follow similar color patterns. A common approach to outline colorization is to use the U-net for color prediction, then further boost performance by employing adversarial training mechanisms~\cite{frans2017outline,hensman2017cgan,zhang2017style} for realistic or artistic colorization.  

 
 
\paragraph{Interactive colorization}
Colorization is essentially a multimodal task in which the desired outcome could vary by person~\cite{charpiat2008automatic}. Numerous studies have introduced new interactive colorization methods to incorporate user color perception~\cite{iizuka2016let,zhang2016colorful, zhang2017real}. These models postulate that user priors are necessary components for real-time user experience, and allow human interaction via global control~\cite{cho2017palettenet,iizuka2016let} or local control~\cite{li2015image,zhang2017real}. However, in most models users generally have limited access during the colorization procedure, due to confusion about how influential a single provided hint will be in the result image. Our model explicitly gives a region of influence, making it easy to understand how the hints provided by the user will be applied. 

\begin{figure}[t]
	\centering
    \includegraphics[width=\textwidth]{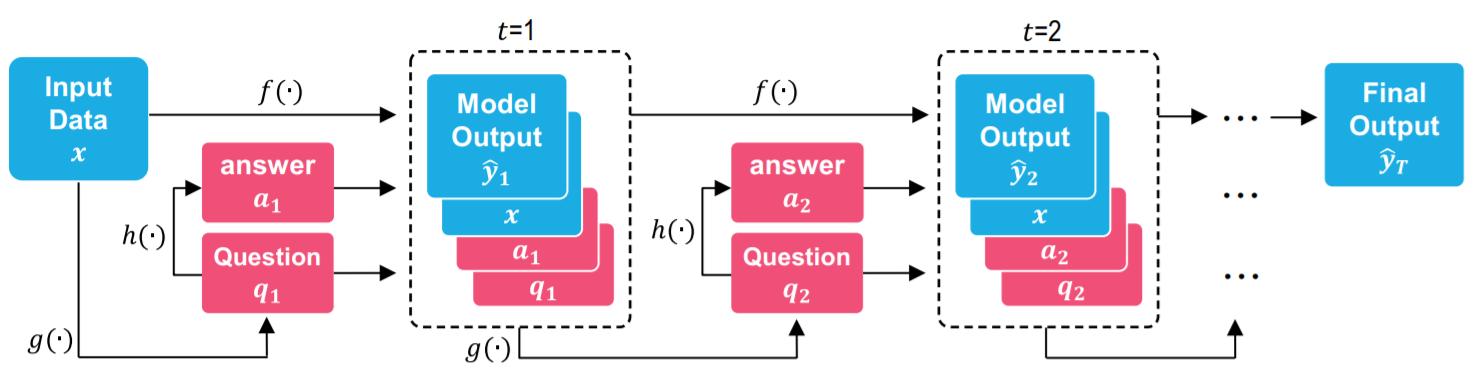}
    \caption{Overview of the proposed asking paradigm. See Section~\ref{sec:asking paradigm} for details.}
    \label{fig:asking_paradigm}
\end{figure}

\section{Asking Paradigm}
\label{sec:asking paradigm}

Given a feature space $\mathcal{X}=\{x_1, x_2, \cdots, x_n\}$ and a target space $\mathcal{Y}=\{y_1, y_2, \cdots, y_n\}$, a conventional approach for supervised learning consists of a predictive function $f(\cdot)$, which models $P(y|x)$ for $y\in\mathcal{Y}$ and $x\in\mathcal{X}$. We now establish our asking paradigm by setting an additional question space $\mathcal{Q}=\{q_1, q_2, \cdots, q_t\}$ and an answer space $\mathcal{A}=\{a_1, a_2, \cdots, a_t\}$. Two new functions $g(\cdot)$ and $h(\cdot)$ map $\mathcal{X}$ to $\mathcal{Q}$ and $\mathcal{Q}$ to $\mathcal{A}$, respectively. The high-level goal is to learn $P(\mathcal{Q}|\mathcal{X})$, so as to optimize $P(\mathcal{Y}|\mathcal{X},\mathcal{Q},\mathcal{A})$. As we assume sequential learning and training of the model, we have the three functions in our framework as follows:
\begin{align} \label{eq:asking paradigm}
\hat{y}_t &= f(x, a_{t-1}, \hat{y}_{t-1}, q_1, q_2, \cdots, q_{t-1})\\
a_t &= h(q_{t}, y)\\
q_t &= g(x, \hat{y}_{t-1}, q_{t-1})
\end{align}
where $t\in{1,2,\cdots,T}$ denotes the timestep for each data instance, $\hat{y}$ is the predicted target, and $h(\cdot)$ is a pre-specified function to provide a hint from the target. Under this framework we attempt to allow the learning agent model both $f$ and $g$. More concretely, we make a model that is provided ($T-1$) opportunities to ask $q_1, \cdots, q_{T-1}$. Answers to the corresponding questions are calculated in the form of the function $a_t = h(q_t, y)$, which is an information-reducing function that provides some limited information about $y$ that corresponds to $q_t$. The calculated $a_t$ is relayed back to the model. The model then produces intermediate predictions $\hat{y}_{1}, \cdots, \hat{y}_{T-1}$ before producing the final output $\hat{y}_{T}$. Loss is then applied based on a differentiable similarity metric between $\hat{y}_{T}$ and $y$. This process encourages models to extract meaningful questions about the final answer. We call this class of networks that model both $f$ and $g$ \textit{asking networks}.

Fig.~\ref{fig:asking_paradigm} shows a generic overview of asking networks. The agent takes as input $x$ and generates a question $q_t$ for every timestep $t$. Based on the question and a pre-defined answer-generating function $h(\cdot)$, the agent obtains an answer to its question. At the next timestep, using all available information the agent predicts the next question and output. The same pattern iterates until the agent receives the error signal at the terminal step $T$. 

\begin{figure}[t]
	\centering
    \includegraphics[width=\textwidth]{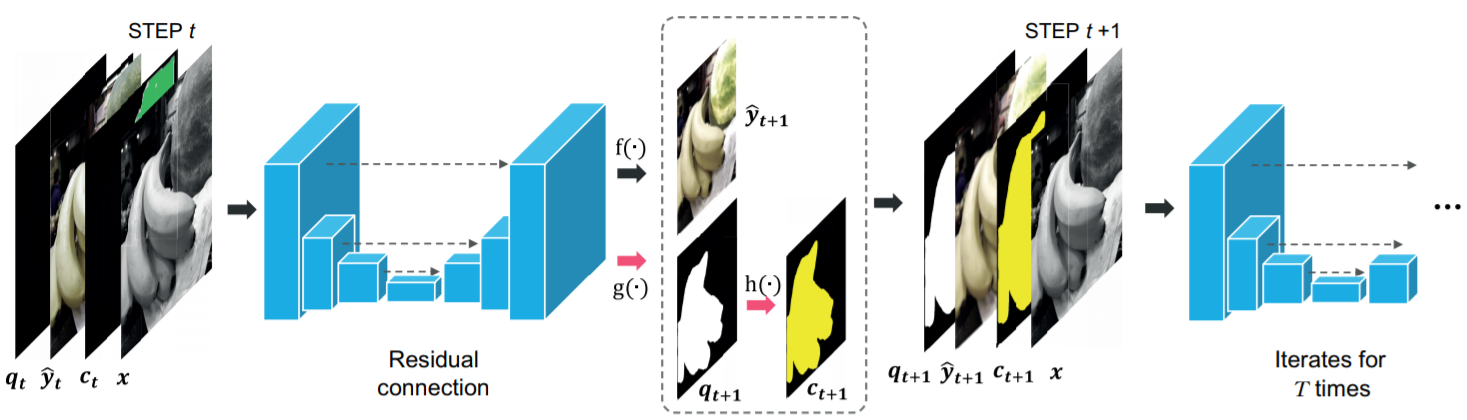}
    \caption{Architecture for deep colorization by asking networks. Inputs at step $t$, $q_t$, $\hat{y}_{t}$, $c_t$ and $x$ are passed to the asking networks, yielding $\hat{y}_{t+1}$ and $q_{t+1}$. We compute $c_{t+1}$ based on a pre-specified rule $h(\cdot)$. All the intermediate outputs are used in the next prediction stage at timestep $t+1$.}
    \label{fig:asking network}
\end{figure}

\section{Automatic Colorization via Asking}
\label{sec:i2i_model}
As a particular application of the above-described asking paradigm, this paper mainly considers an automatic colorization task, which is an example of image-to-image translation where each pixel value of a target image has to be predicted. This task is exemplar of a one-to-many problem~\cite{zhang2017real} that inherently involves ambiguity in prediction tasks. 

A natural, straightforward form of a question in this task would be, say, "what is the (groundtruth) color of a particular sub-region of a target image?". In order to avoid the trivial question of asking the color of the entire region, we limit the provided answer as the \textit{single} color that is the average color of those pixels contained in the sub-region. 

Relaxing the sub-region in a way that each pixel can be partially contained in it, the question $q_t$ has the form of an image-sized heatmap where the range of each pixel value is in $[0, 1]$. Here, the value of 1 indicates the corresponding pixel is completely contained in the sub-region of interest while 0 indicates it is not contained at all. Accordingly, the answer-providing function $h$ is straightforwardly computed as 
\begin{equation} \label{eq:hint_calc}
	\frac{\sum_{i=0}^{H}\sum_{j=0}^{W}\left(q_{t}\odot C\left(y\right)\right)_{ij}}{\sum_{i=0}^{H}\sum_{j=0}^{W}\left(q_{t}\right)_{ij}}
\end{equation}
where $C(\cdot)$ is the characteristic function that calculates the desired local characteristic, say, groundtruth pixel values, of the target image $y$, and $H$ and $W$ are the height and width of the image, respectively. 

As explained in Section~\ref{sec:asking paradigm}, we allow the model to ask questions multiple times in a sequential manner to gradually improve its prediction result of the target $y$ using the provided answers $a_t$s. Specifically, at time $t$, we obtain the predicted target image $\hat{y}_{t+1}$ and $q_{t+1}$ at each timestep, i.e., 
\begin{equation}
\hat{y}_{t+1}, q_{t+1} = f(x, q_t, c_t, \hat{y}_t) \label{eq:model_func}. 
\end{equation}
We transform $a_t$ into the image size by multiplying $q_t$, resulting in a new image $c_t$ as 
\begin{align}
c_t &= q_t \odot a_t \label{eq:answ_calc}\\
\hat{y}_{t+1}, q_{t+1} &= f(x, q_t, c_t, \hat{y}_t). \label{eq:model_func}
\end{align}
Then all images $x, q_t, c_t, \hat{y}_t$ are concatenated to form an input to the network in the next timestep. 
When generating $\hat{y}_1$ and $q_1$, as there is no $\hat{y}_0, q_0,$ or $a_0$, we simply set these inputs with zeros.

When applying Eq.~\ref{eq:asking paradigm} in practice, we do not directly provide all previous hints to the model. Instead, we expect the model to embed and transmit its previous hints through $\hat{y}$. This ensures that the answers provided to the questions of the network are immediately applied. Fig.~\ref{fig:asking network} displays visual illustration of the proposed colorization asking networks. 

The model is given a random integer number of question opportunities $n_{hint} \sim $Unif$(0, T)$ during training. After the model uses a given number of its question opportunities, we apply L2 loss to the final predicted target domain image according to Eq.~\ref{eq:reg_loss}. We also apply a small `smoothing loss' according to Eq.~\ref{eq:smooth_loss} to make questions more understandable to humans; without this smoothing loss, questions tend to be discontinuous. We optimize the model by back-propagating the total loss defined as
\begin{align}  
L_{total} &= L_{reg} + \lambda_{seg}L_{seg} \label{eq:total_loss} \mbox{, where}\\
L_{reg} &= \frac{1}{HW}\sum_{i=0}^{H}\sum_{j=0}^{W} (\hat{y} - y)^2 \label{eq:reg_loss}\\
L_{seg} &= \frac{1}{HW}\sum_{t=0}^{T}\sum_{i=1}^{H}\sum_{j=1}^{W} 
\{|q_t[i][j] - q_t[i-1][j]| + |q_t[i][j] - q_t[i][j-1]|\} \label{eq:smooth_loss}
\end{align} 



\subsection{Improving Question Quality}
\label{sec: making better questions}
During our colorization experiments, we find a few techniques useful for improving the quality of questions generated by asking networks, as follows. 
\paragraph{Smoothing loss} In our objective function defined as Eq.~\ref{eq:total_loss}, if the model happens to optimize only $L_{reg}$ and not $L_{seg}$ while it learns to utilize questions, the generated questions tend to be discontinuous, leaving the questions difficult for humans to interpret. To mitigate this problem, we may apply $L_{seg}$, which encourages the model to have nearby pixels to hold similar values in the question. 
\paragraph{Injecting random noise to answer $a_t$} \label{sec:noisy_answer} The generated questions often exhibit low contrast between those pixels mainly contained in them and those not. In other words, the model essentially seeks for the color answer across multiple different objects, reducing the training efficiency. To address the problem, we inject a small amount of random noise to $a_t$. Such random noise makes small color difference indiscriminable, thus enforcing the model to learn to
identify colors and objects as clearly as possible. We observe that it learns to suppress the regions that it does not highlight and focuses its question solely on a relatively small region, which mostly corresponds to a single object. A significant performance gain from the noise-injecting technique is reported in Section~\ref{sec: class precision}.

\subsection{Peripheral Details}
\label{sec:training_settings}
We use a U-net~\cite{ronneberger2015u} as the network architecture of our model. In addition to the main output channels that contain the predicted color values of an uncolored input image, our output contains an additional channel that acts as a question heatmap generated by a sigmoid layer at the end, rendering the values ranging from 0 to 1. We use $T=4$ for natural images and $T=6$ for cartoon images. We use the Adam optimizer~\cite{DBLP:journals/corr/KingmaB14} to optimize the neural networks. When we colorize real-world images, we use the CIELAB color space and try to predict the a*b* channel value given the grayscale L channel. When we colorize comic images given an outline sketch of images, we predict the RGB channel values. We use the 2011 ImageNet training dataset~\cite{russakovsky2015imagenet} to train the real-world image colorization models. For the cartoon images, a set of images collected from the NAVER WEBTOON platform is used; they are used with the authorization of the artists, which was obtained with the support of NAVER WEBTOON Corp. 

\section{Experiments}
\subsection{Quantitative Analysis}


\begin{figure}[t]
\centering
\includegraphics[width=\textwidth]{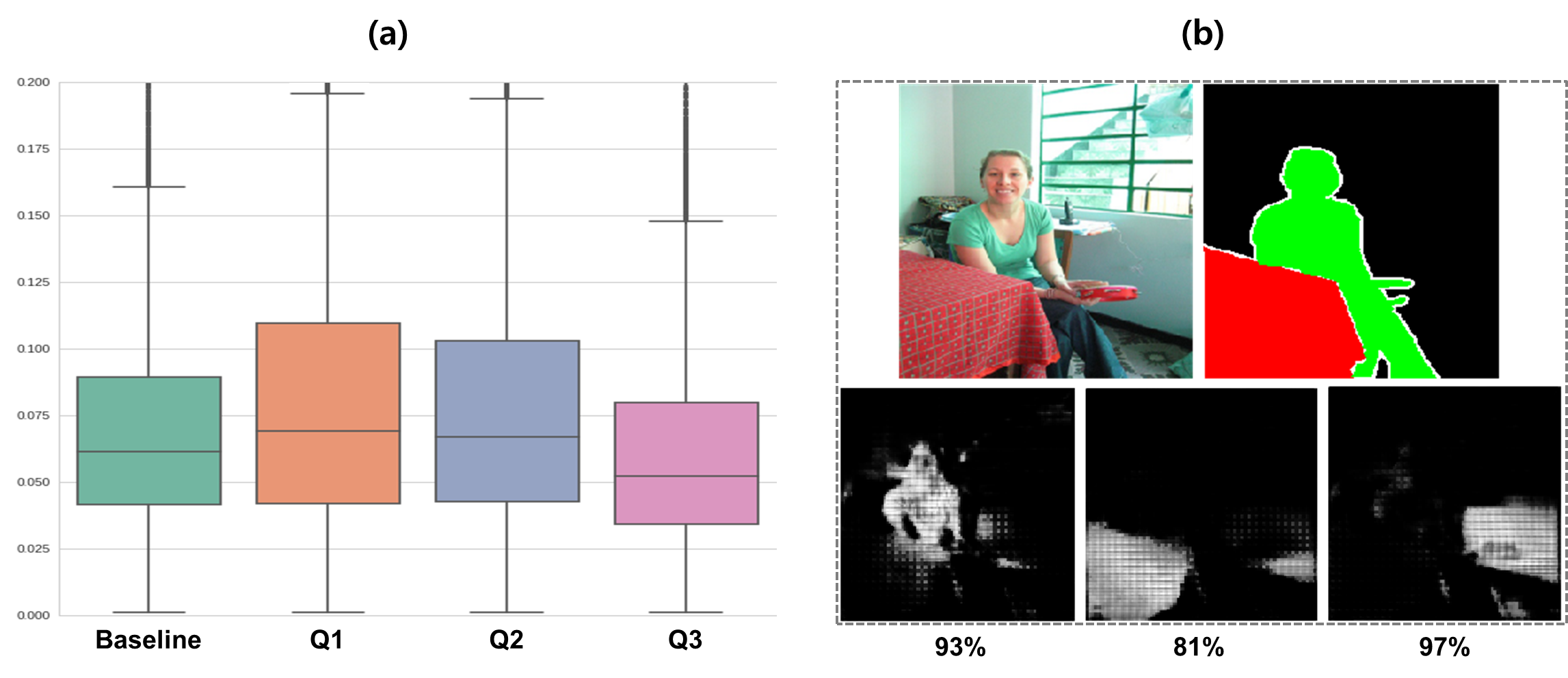}
\caption{\textbf{(a)} Order of questions; the model asks challenging regions first. See Section~\ref{sec:order_of_hints} for details. \textbf{(b)} Visual aid for understanding class precision rate. Precision rate measures how much proportion of question overlap with a single segmented class region. Sample measures are written under each question heatmap.}
\label{fig:order_hint}
\end{figure}

\subsubsection{Effects of Question Order}
\label{sec:order_of_hints}
 We allow the model to ask several questions per image. However, the number of questions varies randomly from 0 to $T$ times during training. Thus the model does not know in advance how many chances will be given. It drives the model to take greedy actions to maximally benefit from every question opportunity. To evaluate how much the model benefits from each question, we train an auxiliary colorization network following the main network architecture proposed in~\citet{zhang2017real} as a baseline, trained with no hints at all. We use the baseline to measure the average loss per every pixel of an image, and whether our model actually asks those regions ambiguous to colorize. 
 To this end, using 10,000 ImageNet validation images, we forward-propagate each image using the baseline model to computing the pixel-wise $L_1$ error map $E\in\mathbb{R}^{W\times H}$ in the color prediction result. Using the same input image, we generate three questions from the trained asking network of our model and obtain the three question maps $Q_i$'s for $i=1, 2, 3$. 
Afterwards, we compute the global average error per pixel as $\sum_{i=0}^{H}\sum_{j=0}^{W}E_{ij}$, which we call a baseline, as well as its weighted version using $Q_i$ as the weights, i.e., $\sum_{i=0}^{H}\sum_{j=0}^{W}\left(E\odot Q_{i}\right)_{ij}$ for $i=1, 2, 3$. 
 
Fig.~\ref{fig:order_hint}(a) shows the distributions of these errors across 10,000 validation images.
The highest errors of Q1 among the baseline as well as all the questions indicate that the first question asks about the color of regions that the model predicts worse than any other regions. Q2 asks about those regions that the model can predict slightly better than Q1 but still worse than the global average. Finally, Q3 asks about those regions with substantially low prediction error. 
These results imply that our model learns to greedily ask what it finds as the most challenging to predict at every step.

\begin{table}[t]
\begin{center}

\begin{tabular}{ccccccc}  
\toprule
Method &Steps&
\begin{tabular}[c]{@{}c@{}}\textbf{Endo et al.}\\~\cite{endo2016deepprop}\end{tabular}&
\begin{tabular}[c]{@{}c@{}}\textbf{Barron and}\\\textbf{Poole}~\cite{barron2016fast}\end{tabular} &
\begin{tabular}[c]{@{}c@{}}\textbf{Levin et al.}\\~\cite{levin2004colorization}\end{tabular}&
\begin{tabular}[c]{@{}c@{}}\textbf{Zhang et al.}\\~\cite{zhang2017real}\end{tabular}&\textbf{Ours} \\ 


\midrule
\multirow{4}{*}{Rand}&0 &22.82 $\pm$ 0.52 & 22.82 $\pm$ 0.30 &22.82 $\pm$ 0.30 & $24.43$ $\pm$  0.14& $\mathbf{24.64}$ $\pm$ 0.13\\
&1 &22.96 $\pm$ 0.51 & 23.28 $\pm$ 0.30 &24.01 $\pm$ 0.30 & 25.37 $\pm$  0.14& $\mathbf{28.16}$ $\pm$ 0.14\\
&2 &22.93 $\pm$ 0.52 & 23.55 $\pm$ 0.29 &24.85 $\pm$ 0.29 & 26.19 $\pm$  0.14& $\mathbf{29.32}$ $\pm$ 0.15\\
&3 &23.44 $\pm$ 0.51 & 23.85 $\pm$ 0.29 &25.27 $\pm$ 0.29 & 26.69 $\pm$  0.14& $\mathbf{29.62}$ $\pm$ 0.14\\
\midrule
\multirow{4}{*}{Max}
&0 &23.13 $\pm$ 0.31 & 22.97 $\pm$ 0.30 &22.97 $\pm$ 0.30 & 24.43 $\pm$  0.14& -\\
&1 &18.21 $\pm$ 0.27 & 23.82 $\pm$ 0.30 &19.22 $\pm$ 0.29 & 25.58 $\pm$  0.14& -\\
&2 &20.95 $\pm$ 0.29 & 24.54 $\pm$ 0.29 &23.68 $\pm$ 0.29 & 26.94 $\pm$  0.14& -\\
&3 &22.14 $\pm$ 0.29 & 25.04 $\pm$ 0.29 &24.94 $\pm$ 0.28 & 27.76 $\pm$  0.14& -\\
\bottomrule
\end{tabular}
 \\[0.9em]
\caption{\label{table:PSNR}\textbf{PSNR Comparisons}; we compare performances of our model with scores of the other four interactive colorization baselines.}
\end{center} 
\end{table}
 
\subsubsection{Performance Comparisons of Hint-based Colorization}
In this experiment, we analyze the gain per answer given to the question asked by our model, in the context of interactive colorization. To this end, we compare the performance gain in terms of the peak signal-to-noise ratio (PSNR) between our proposed method and other hint-based colorization models~\cite{barron2016fast,endo2016deepprop,levin2004colorization,zhang2017real}. Specifically, in our model, we calculate the performance gain per every answer given to the model-generated question. In other baseline models, we follow \citet{zhang2017real} and adopt two methods to compute performance, which are hints (or groundtruth colors) at random positions (Rand) and at the positions that has the highest errors in $E$ (Max). 


Table~\ref{table:PSNR} shows the results of this experiment. Compared to the results of Rand methods, our model outperforms all the baseline models at all of the steps. It is notable that our model performs better than any other baseline with only one hint. We attribute the competitive performance to the early question regions being the most difficult regions to predict. Qualitative results (see Section~\ref{Sec:qualitative analysis}) suggest our model learns to distinguish semantically meaningful objects based on color similarity and then ask about the region belonging to a single object with consistent colors. In other words, our asking network reveals that the colorization model implicitly learns object segmentation in an intelligent manner while learning to colorize.

\subsubsection{Class Precision Analysis with VOC Segmentation Dataset}
\label{sec: class precision}
To test whether the region corresponding to a particular question is semantically meaningful, we measure the precision of heatmaps to be within the same class object in an image. To this end, we use Visual Object Classes Challenge 2012~\cite{pascal-voc-2012} (VOC2012) segmentation dataset to generate four questions on a model trained on ImageNet. In detail, we match each question heatmap with 22 different classes of VOC2012 dataset to find the class with the highest match. To illustrate, in Fig.~\ref{fig:order_hint} (b), three question maps are shown at the bottom, and we compute in what percentage each question overlaps (precision) with the class segmentation map shown on the top-right position. For instance, the first question records 93\% of precision with the \textit{person} class. By averaging the precision values across all VOC2012 images and question maps we compute the overall precision.

We record the precision performance of 75.4\% without the noise injection technique (Section~\ref{sec:noisy_answer}). As discussed in Section~\ref{sec:noisy_answer}, adding random noise to the answer $a_t$ provides improved results both qualitatively and quantitatively. The questions shown in Fig.~\ref{fig:order_hint} (b) display high contrast between pixels, and it clearly shows which region the model is asking. In addition, we record a precision rate of 86.7\% with the random noise, which suggests that our model asks about the region corresponding mostly to a single object. 

\subsection{Qualitative Analysis}
\label{Sec:qualitative analysis}
\begin{figure}[h]
	\centering
    \includegraphics[width=\textwidth]{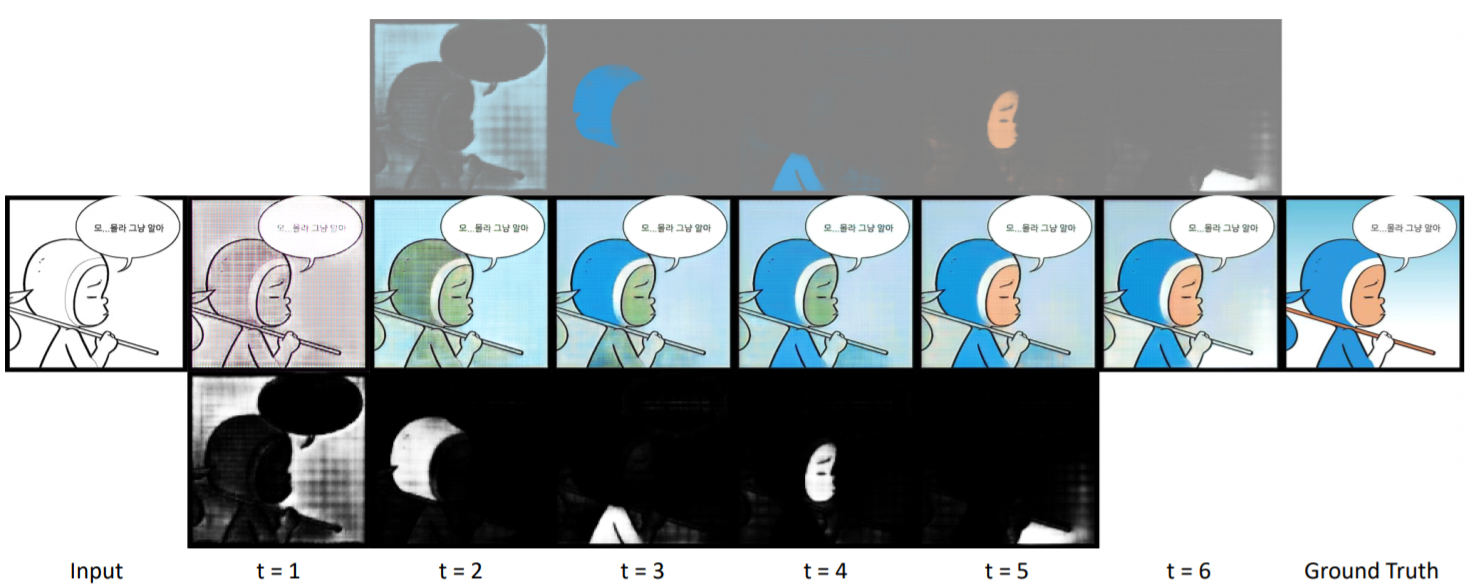}
    \caption{Visualization of how an outline drawing image becomes colorized step by step. \copyright Donggeon Lee.}
    \label{yumi_main_img}
\end{figure}

In this section, we visually illustrate the question maps and the step-by-step colorization process of our model. To show that our model works well without grayscale input which may act as a hint for segmentation, we also train another model using the comic images datasets described in Section~\ref{sec:training_settings} and show colorized results on test images. Our model generates distinctive segmentations given simple outlines. More examples can be found in the supplemental material. Figs.~\ref{img_itr_color01} and \ref{yumi_main_img} show the example colorization results of a natural image and a cartoon image, respectively.  The images in the middle row are intermediate colorization results, the bottom row shows questions generated by the model while the top row images represent answers provided. One can see that the model asks semantically coherent questions. 

\begin{figure}[h]
	\centering
    \includegraphics[width=\textwidth]{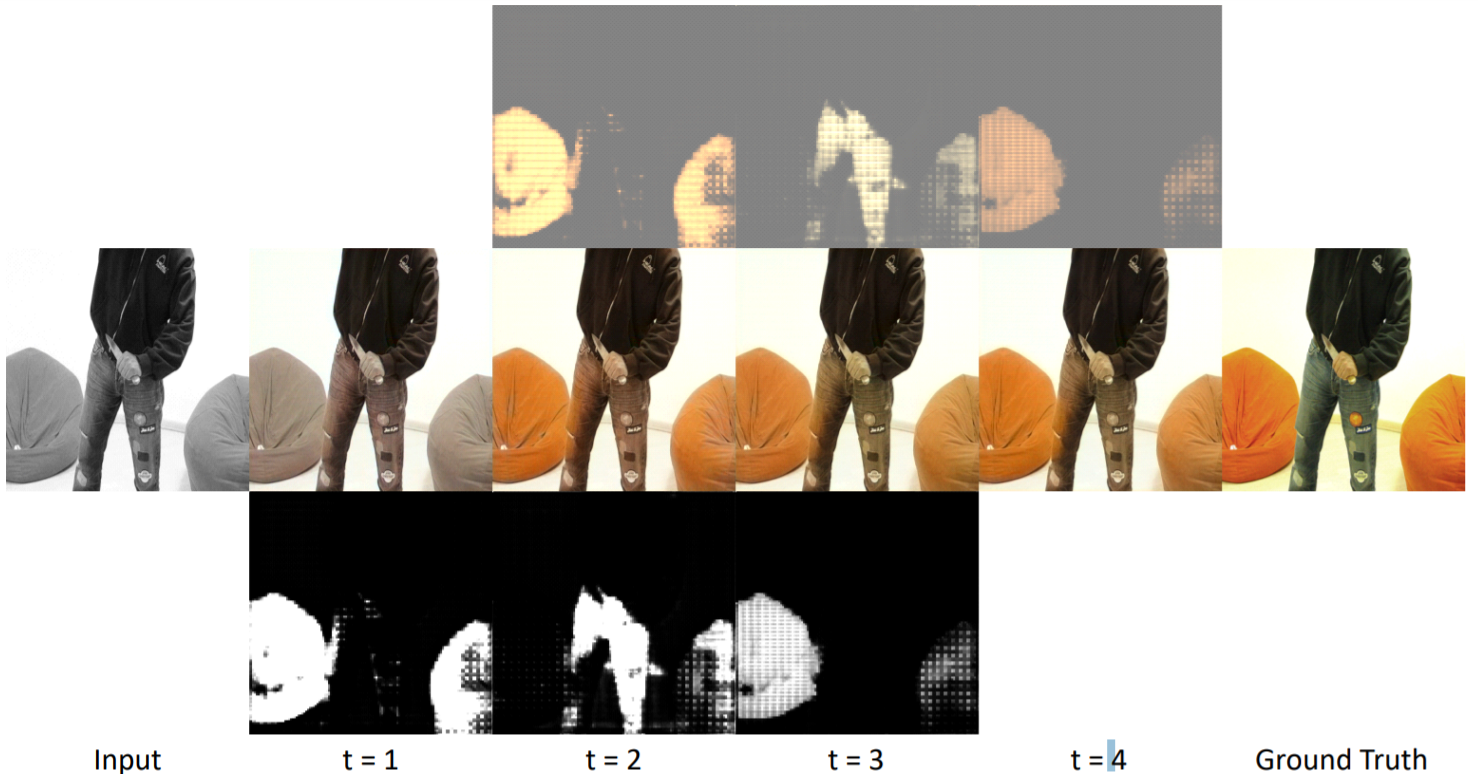}
    \caption{Example colorization process of a natural image.}
    \label{img_itr_color01}
\end{figure}

\section{Conclusion and Future Work}
Questioning is an effective source of learning for human beings, and it is also an essential way of telling what a person knows and does not. Thus providing a network the opportunity to question, so as to let it reveal by itself its weaknesses is an important milestone in machine learning. In this work, we proposed a general framework in which models are allowed to ask about parts of the answer, and suggested a novel approach in image colorization. Both quantitative and qualitative analyses show that the proposed colorization model creates meaningful questions readily interpretable and understandable by humans. This poses as a promising new approach for interpretability of neural networks, in which they do not only produce results, but can actively exhibit \textit{why} they thought in such a manner, due to which part of the answer. 

Nonetheless, our model has room to improve. A problem is that training takes a relatively long time; it requires a significant amount of training time until the model learns to properly generate different questions at different timesteps. 
Our future work involves building theoretical foundation the optimal training of the asking networks. 

Nevertheless, we believe that the proposed asking networks provide multiple interesting research directions. One bright side of our model is that it naturally enables an interactive steering of deep neural networks. Since the model asks questions in a human-readable or an interpretable form, we could communicate with the machine and steer the final output production to what we desire, via any custom answers. We briefly show how an interactive steering of deep colorization works in the supplemental material. 

Based on the asking framework, it is possible to improve the interpretability of other image-to-image translation tasks, such as inpainting~\cite{pathak2016context} or expression transfer~\cite{thies2015real}. Moreover, we do not limit the scope of potential application areas within computer vision, but some other domains such as natural language processing could also largely benefit from the proposed asking paradigm to make outcomes interpretable. For instance, in machine translation~\cite{bahdanau2014neural}, one might adopt the asking networks in the decoding stage to find out in which part of the word sequence prediction the model strives most. Our future work therefore involves extending our asking paradigm to a broader range of domains including computer vision as well as natural language processing.  

\subsection*{Acknowledgments}
We would like to express our gratitude towards NAVER WEBTOON Corp. and the comic artists Donggeon Lee, Pipp Choi, Tae Hoon Shin /Seung Hoon Ra, Joong Rok Kook/Sang Sin Lee, and Omyo for providing the comic images for this research. 



\bibliographystyle{plainnat}
\bibliography{nips_bib}

\medskip

\small

\end{document}